\documentclass[letterpaper]{article} % DO NOT CHANGE THIS
\usepackage{aaai23}  % DO NOT CHANGE THIS
\usepackage{times}  % DO NOT CHANGE THIS
\usepackage{helvet}  % DO NOT CHANGE THIS
\usepackage{courier}  % DO NOT CHANGE THIS
\usepackage[hyphens]{url}  % DO NOT CHANGE THIS
\usepackage{graphicx} % DO NOT CHANGE THIS
\usepackage{natbib}  % DO NOT CHANGE THIS AND DO NOT ADD ANY OPTIONS TO IT
\usepackage{caption} % DO NOT CHANGE THIS AND DO NOT ADD ANY OPTIONS TO IT
\usepackage{algorithm}
\usepackage{algorithmic}

\usepackage{newfloat}
\usepackage{listings}
\DeclareCaptionStyle{ruled}{labelfont=normalfont,labelsep=colon,strut=off} % DO NOT CHANGE THIS
\floatstyle{ruled}
\newfloat{listing}{tb}{lst}{}
\floatname{listing}{Listing}
\usepackage[backref=page,pdfpagemode=FullScreen,colorlinks=true,pagebackref=true]{hyperref} % -- This package is specifically forbidden

\usepackage{amsmath}
\usepackage{amssymb}
\usepackage{booktabs}
\usepackage{comment}
\usepackage{subcaption}
\usepackage{colortbl}
\usepackage{lscape}
\usepackage{xcolor}
\usepackage{tabularx}
\usepackage{multirow,bigstrut}
\usepackage{nicefrac}
\usepackage{adjustbox}
\usepackage{diagbox}

\newcommand{\methodname}{DILEMMA}
\newcommand{\methodnameexpansion}{Detection of Incorrect Location EMbeddings with MAsked inputs}
\newcommand{\real}{\mathbb{R}}
\newcommand*\rot{\rotatebox{90}}
\title{Representation Learning by Detecting Incorrect Location Embeddings}
\author {
    Sepehr Sameni\textsuperscript{\rm 1},
    Simon Jenni\textsuperscript{\rm 2},
    Paolo Favaro\textsuperscript{\rm 1}
}
\affiliations {
    \textsuperscript{\rm 1} Computer Vision Group, University of Bern, Switzerland\\
    \textsuperscript{\rm 2} Adobe Research\\
    sepehr.sameni@unibe.ch, jenni@adobe.com, paolo.favaro@unibe.ch
}

\begin{document}

\maketitle

\begin{abstract}
In this paper, we introduce a novel self-supervised learning (SSL) loss for image representation learning. There is a growing belief that generalization in deep neural networks is linked to their ability to discriminate object shapes. Since object shape is related to the location of its parts, we propose to detect those that have been artificially misplaced. We represent object parts with image tokens and train a ViT to detect which token has been combined with an incorrect positional embedding. We then introduce sparsity in the inputs to make the model more robust to occlusions and to speed up the training. We call our method \methodname, which stands for \methodnameexpansion. We apply {\methodname} to MoCoV3, DINO and SimCLR and show an improvement in their performance of respectively $4.41$\%, $3.97$\%, and $0.5$\% under the same training time and with a linear probing transfer on ImageNet-1K. We also show full fine-tuning improvements of MAE combined with our method on ImageNet-100. We evaluate our method via fine-tuning on common SSL benchmarks. Moreover, we show that when downstream tasks are strongly reliant on shape (such as in the YOGA-82 pose dataset), our pre-trained features yield a significant gain over prior work.\footnote{source code: \url{https://github.com/Separius/DILEMMA}}

\end{abstract}

\section{Introduction}

In computer vision, deep learning models trained on small labeled datasets can benefit greatly from supervised pre-training on datasets such as ImageNet \cite{girshick2014rich}. Even more surprisingly, \cite{he2020momentum} showed that it is possible to pre-train with unlabeled data (with MoCo) and outperform pre-training with supervised learning on several downstream tasks. This led to the rapid development of several Self-Supervised Learning (SSL) methods, such as \cite{caron2020unsupervised,chen2020simple,chen2020improved,he2020momentum}.

Representations obtained via SSL have the ability to \emph{generalize} to downstream tasks such as object classification, detection, and segmentation \cite{deng2009imagenet,Everingham2009ThePV,Krizhevsky2009LearningML}. Recent work suggests that representations with a \emph{shape bias} generalize better to these tasks than those with a texture bias \cite{geirhos2018imagenet,tartaglini2022developmentally}.

\begin{figure}[t]
	\centering
    \newlength{\yfh} % yoga figure height
    \setlength{\yfh}{0.1\textwidth}
    \newlength{\yfw} % yoga figure width
    \setlength{\yfw}{0.115\textwidth}
    \begin{adjustbox}{width=\columnwidth}
    \begin{tabular}{cccc}
	    \includegraphics[height=\yfh,width=\yfw]{"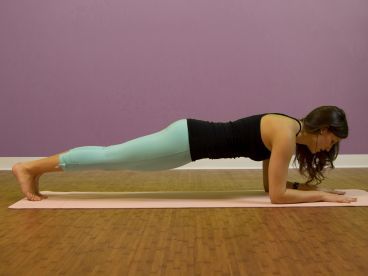"} &
	    \includegraphics[height=\yfh,width=\yfw]{"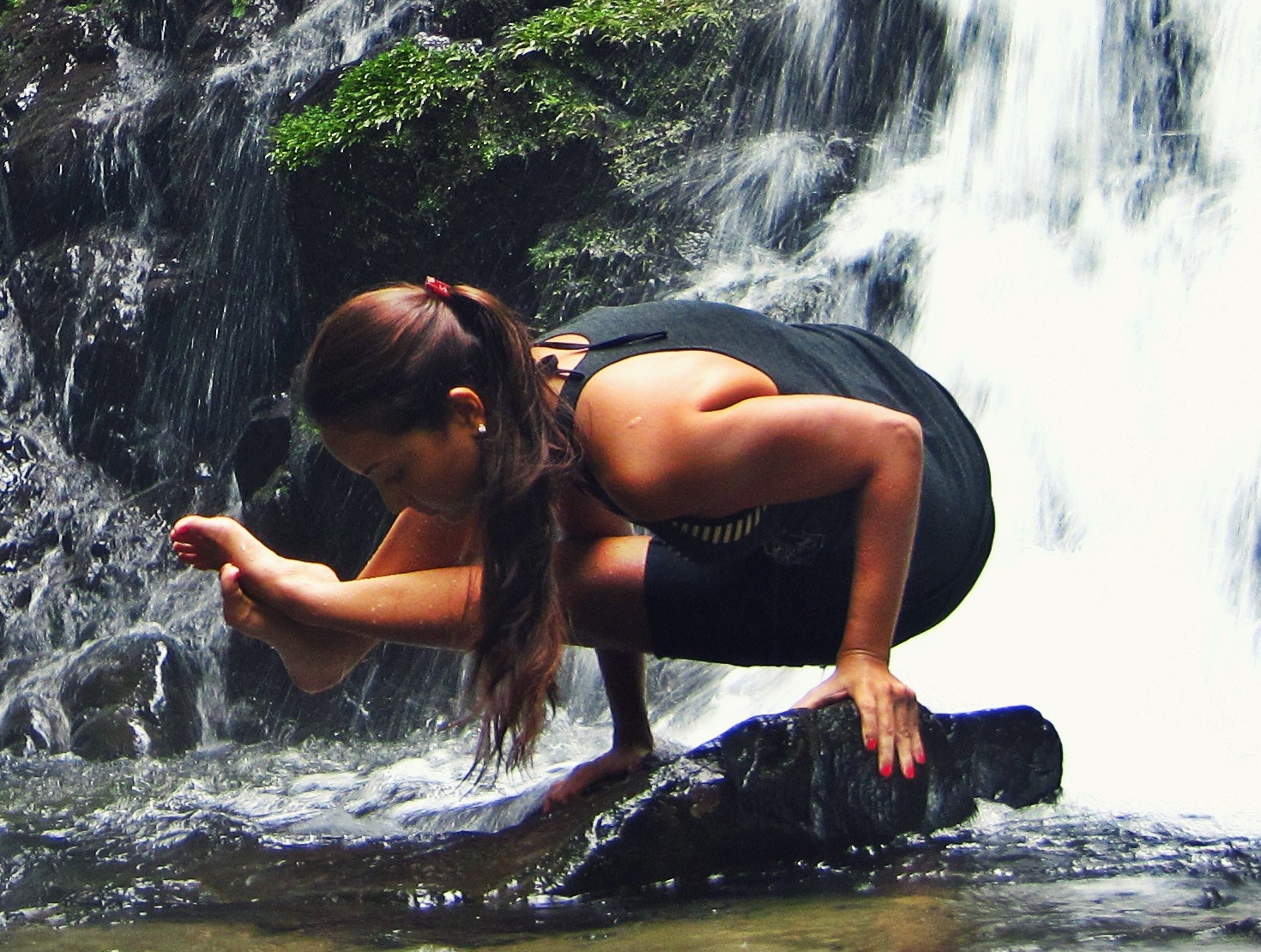"} &
		\includegraphics[height=\yfh,width=\yfw]{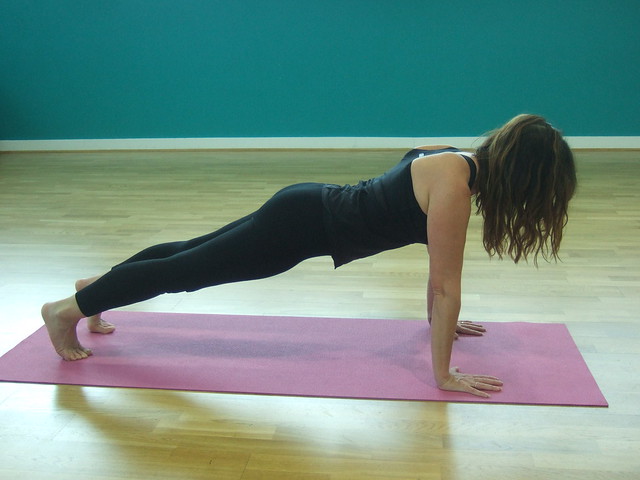} &
		\includegraphics[height=\yfh,width=\yfw]{"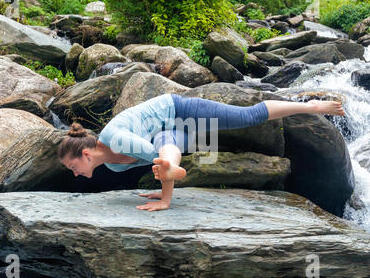"} \\
        \footnotesize{Dolphin plank} & \footnotesize{Eight angle} & \footnotesize{Plank} & \footnotesize{Sage Koundinya}
    \end{tabular}
    \end{adjustbox}
	\caption{$\text{Yoga}_{82}$ is a dataset with a shape-based task. As these examples show, texture alone is not sufficiently indicative of the pose. Thus, models that perform well in this task may demonstrate a strong shape discriminability.}
	\label{fig:yoga82}
\end{figure}

In particular, the expectation is that image representations may do better in the transfer learning to a shape-based task, such as pose classification of  $\text{Yoga}_{82}$ (see Fig.~\ref{fig:yoga82}). Thus, we investigate whether adding a regularization loss that is sensitive to shape to a state-of-the-art SSL method, might lead to better representation learning.

We propose {\methodname}, which is short for \methodnameexpansion. In our experiments we integrated {\methodname} with SimCLR~\cite{chen2020simple}, DINO~\cite{caron2021emerging} and MoCoV3~\cite{chen2021empirical}, and find a consistent improvement across the majority of downstream tasks.

With \methodname, the image representation is encouraged to differentiate shapes thanks to two main components: 1) A binary classification loss to detect the correct/incorrect positions of object parts, and 2) the use of randomized input sparsity, so that every subset of object parts contributes to the whole image representation.
The first component is a concept already proposed in the context prediction \cite{doersch2015unsupervised} and jigsaw puzzle SSL methods \cite{noroozi2016unsupervised}.
It takes also inspiration from ELECTRA \cite{clark2020electra}, where some text tokens are replaced by a weak generator and a discriminator is trained to detect them. The second component, is also a concept that has been exploited in 
VATT~\cite{akbari2021vatt} and MAE~\cite{he2021masked} to reduce the computational workload of training with ViTs \cite{dosovitskiy2020image}.

More in detail, as shown in Fig.~\ref{fig:DILEMMA}, we split an image into a grid of tiles, map them to tokens, and combine them with positional embeddings. Then, we corrupt the positional embeddings of a fraction of the tokens before we feed them to a ViT. In our {\methodname} loss, we classify the tokens into those with correct and incorrect positional embeddings. 
The \emph{sparsification} of the input can be implemented in a ViT simply by discarding a randomized percentage of tokens.

We also use a teacher-student architecture as in MoCoV3 \cite{chen2021empirical}. We sparsify only the input to the student network and instead feed all the tiles to the teacher network. Because it is used only in evaluation mode, it does not have a significant impact on storage and computing resources. Moreover, the use of a complete set of tiles (a setting that we call \emph{dense}) and without corrupted positional embeddings, allows the teacher to build a better reference for the student network.

\noindent\textbf{Our contributions can be summarized as follows:}
\begin{itemize}
\item We introduce {\methodname}, a novel SSL regularization loss that enhances the shape discriminability of image representations; it is based on the detection of misplaced positional embeddings with a ViT and the use of sparsity in the input;
\item We propose to randomly sparsify the inputs and to use a student-teacher architecture to: 1) reduce the memory storage, 2) close the gap between training and test data, 3) speed up the training; 
\item {\methodname} boosts the performance of MoCoV3, SimCLR, DINO, and MAE under the same computational budget.
\end{itemize}

\section{Related Work}

\noindent\textbf{Self-Supervised Learning for Image Representations.} Self-supervised learning gained popularity as a form of unsupervised learning, where pretext tasks leverage supervision signals obtained without human labor. 
Some classic examples are the classification of image patch locations \cite{doersch2015unsupervised,noroozi2016unsupervised}, the reconstruction of color channels \cite{zhang2016colorful} or image patches \cite{pathakCVPR16context}, or the recognition of various image transformations  \cite{gidaris2018unsupervised,jenni2018self},
While prior patch-based methods inspired our approach of detecting wrongly placed image patches, ours is both simpler and performs better in transfer experiments. Furthermore, due to the input representation of ViTs (disjoint image patches) and our random sparse patch sampling, our approach suffers less from domain gaps between pre-training and transfer.\\ 
\noindent\textbf{Contrastive Learning.}
Efforts to scale up and improve instance discrimination \cite{dosovitskiy2015discriminative,wu2018unsupervised} as a self-supervised pre-training task have established contrastive learning \cite{chen2020simple,he2020momentum,oord2018representation} as the most popular SSL approach in computer vision today.
Several modifications of the basic recipe, \emph{i.e.}, learning to discriminate training instances up to data augmentations, have been proposed since. 
For example, some methods leverage momentum encoded samples for positive and negative sampling \cite{he2020momentum,chen2020improved}, some remove the need for explicit negative pairs \cite{Grill2020BootstrapYO,chen2020exploring}, and others extend the set of positives beyond data-augmentation through clustering \cite{caron2020unsupervised} or nearest-neighbors in feature space \cite{dwibedi2021little}. 
Another line of work considers contrastive pre-training strategies tailored to dense prediction tasks \cite{o2020unsupervised,wang2021dense,xiao2021region,xie2021propagate,li2021dense,liu2021efficient}. 
More recently, contrastive methods leverage vision transformer architectures \cite{dosovitskiy2020image,liu2021swin}, \emph{e.g.}, by adapting existing approaches \cite{chen2021empirical,xie2021self}, tailoring architectures \cite{li2021efficient}, or novel objectives \cite{caron2021emerging}. 
In our approach, we show that several well-established contrastive baselines \cite{chen2021empirical, caron2021emerging, chen2020simple} can be improved through the addition of a spatial reasoning task and by extending the set of image augmentations through randomized patch dropping. \\
\noindent\textbf{Self-Supervised Pre-Training of Transformers.}
The success of the transformer architecture \cite{vaswani2017attention} in natural language is to a great extent due to large-scale self-supervised pre-training tasks.
Successful pre-training strategies from NLP like masked token prediction \cite{devlin2018bert} have recently also been adapted to the image domain \cite{bao2021beit,zhou2021ibot,he2021masked,zhou2021ibot}. 
Our patch misplacement detection is similar to another type of pretext task in NLP, where the goal is to detect corrupted tokens, \emph{i.e.}, words replaced by an imperfect masked language model \cite{clark2020electra,clark2020pre}.
However, a key difference in our approach is that we only tamper with the spatial position of the tokens and thus do not require a separate masked token prediction model. In parallel work, Fang et al.~\cite{Fang2022CorruptedIM} use BEiT~\cite{bao2021beit} for that purpose.
The method of DABS~\cite{tamkin2021dabs} also uses the idea of patch misplacement, but it does not have a way to handle degenerate learning and it does not show performance improvements. MP3~\cite{Zhai2022PositionPA} also predicts the position of all the tokens like jigsaw~\cite{noroozi2016unsupervised} with a ViT. A technique that has proven very beneficial to improve the training efficiency of vision transformers is token dropping \cite{akbari2021vatt,he2021masked,el2021large,chen2022context}. 
We extend this technique by randomizing the token dropping amount and including the case of no dropping to narrow the domain gap between pre-training and transfer.

\section{Training \methodname}

\begin{figure*}[t]
	\centering
	\includegraphics[width=2\columnwidth]{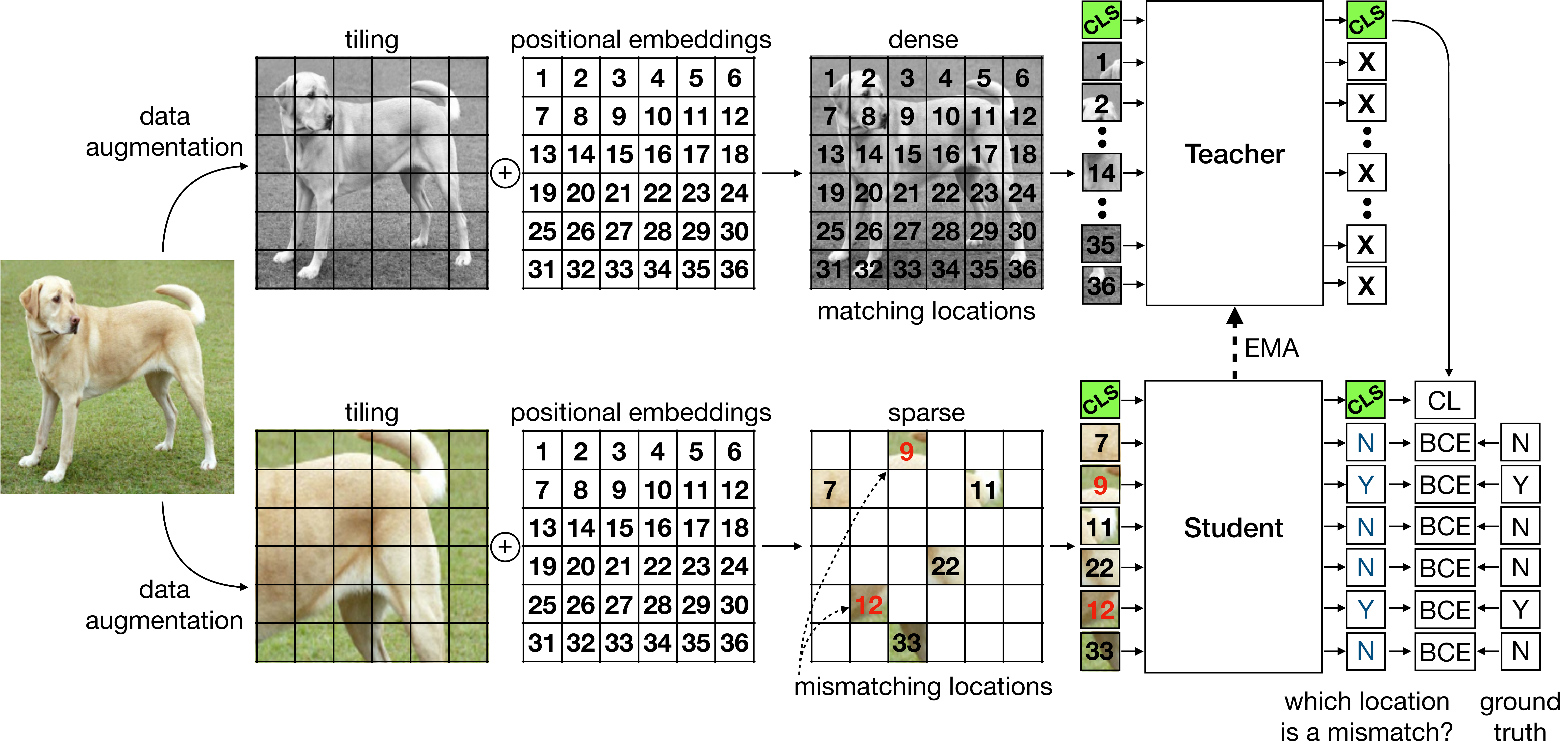}
	\caption{Training with \methodname: A sample image is augmented twice and split into tiles (we use a $14\times 14$ grid). The Teacher network takes the complete set of tiles as input (dense) and without mismatches in the positional embeddings for each token. The Student takes only a subset of the tiles as input (sparse) and some tiles have incorrect positional embeddings. The Student is then trained under two losses: one is the contrastive loss of the class tokens (CLS) between the Teacher and the Student, and the other is the {\methodname} binary cross-entropy for each token.
	}
	\label{fig:DILEMMA}
\end{figure*}

Let us define an image sample as $x\in \real^{H\times W \times C}$, \emph{i.e.}, $x$ has $H\times W$ pixels and $C$ color channels. We apply two data augmentations \cite{Grill2020BootstrapYO} to $x$ and obtain $\hat{x}_1$ and $\hat{x}_2$. Similarly to ViT, each input $\hat{x}_1$ and $\hat{x}_2$ is divided in $14\times 14$ tiles, flattened and projected to $N$ tokens $t_{1,i}, t_{2,i}\in \real^{D}$, $\forall i \in U\doteq \{1,\dots,N\}$, through a linear projection. We then combine each token $t_{\cdot,i}$, with a positional embedding $p_i\in\real^D$, which can be either learned or fixed.

As in MoCoV3 \cite{chen2021empirical}, we define a \emph{Student} $S$ and a \emph{Teacher} $T$ ViTs \cite{dosovitskiy2020image}, where the Teacher, also called \emph{momentum encoder}, is obtained through the exponential moving average (EMA) of the Student's weights (thus, it is not trained). 
The Teacher receives as input all the tokens $t_{1,1},\dots,t_{1,N}$ with the corresponding positional embeddings $p_1,\dots,p_N$.
The Student instead receives as input a sparse set $M\subset U$ of tokens $t_{2,i}$, $i\in M$. For a randomized fraction of these tokens $B\subset M$ the corresponding positional embdeddings $q_{i}$, $i\in M$ are incorrect, \emph{i.e.}, $q_{i} \doteq p_{i}$ if $i\in M\setminus B$ and $q_{i} \doteq p_{j}$ with $j\in U \setminus M$, if $i\in B$. We call the ratio $\theta = \nicefrac{|B|}{|M|}\in [0,1]$, between the cardinalities of $B$ and $M$, the \emph{probability of a positional embedding mismatch}.
We choose a different $M$ and $B$ sets for each sample at each iteration.
We define a set of ground truth labels $y_i=0$ (N) if $i\in M\setminus B$ and $y_i=1$ (Y) if $i\in B$. The $i$-th output token from the Student is denoted with $S_i(\{q_j\oplus t_{2,j}\}_{j\in M})$. We indicate the extra classification token with $i=0$ both at the input and output. Also, $q_{0},p_{0}=0$, \emph{i.e.}, no location encoding.

Now we are ready to introduce the {\methodname} loss (see also the whole training method in Fig.~\ref{fig:DILEMMA})
\begin{multline}
\!\!\!\!\!\!\!\textstyle {\cal L}_\text{\methodname} = \mathbb{E}_x\big[ \textstyle\sum\limits_{i\in M}y_i\log\left(\sigma\left(Y S_i\left(\{q_j\oplus t_{2,j}\}_{j\in M \cup \{0\}}\right)\right)\right) \\ \textstyle+\left(1-y_i\right) \log\left(1-\sigma\left(Y S_i\left(\{q_j\oplus t_{2,j}\}_{j\in M \cup \{0\}}\right)\right)\right)\big]
\label{eq:dilemma}
\end{multline}
where $\mathbb{E}[\cdot]$ is the expectation over image samples, $\sigma$ is the sigmoid function and $Y$ is a linear projection.

Because of the sparsity in the input to the Student network, we also obtain a computational benefit. When we increase the sparsity of the input, we can also increase the mini batch size to fully utilize the GPU RAM. This is particularly significant with ViTs, because of their quadratic scaling with the number of tokens (the memory usage is $O(N^2)$). The fact that we can significantly increase the mini batch size is particularly effective with contrastive learners. Moreover, in this way it is also faster to train our model, because the average mini batch size is much larger than when using dense inputs (in our case it is $2.5\times$ more).

\subsection{Combining {\methodname} and Contrastive Learning}
\label{sec:losses}

The {\methodname} loss can be integrated with other SSL losses. Here we describe the integration with the contrastive loss, but other choices follow an identical procedure.

The contrastive loss is defined as
\begin{multline}
\textstyle {\cal L}_\text{CNT} = \mathbb{E}_x\big[ \textstyle L_\text{CE}\bigl(S_0\left(\{q_j\oplus t_{2,j}\}_{j\in M \cup \{0\}}\right),\\T_0\left(\{p_j\oplus t_{1,j}\}_{j=0,\dots,N}\right)\bigr)\big],
\end{multline}
where
\begin{align}
\textstyle    L_\text{CE}\left(A, V\right) = -2\tau\sum_n z_n\log \text{softmax}\left(\frac{A_n^\top V}{\tau}\right)
\end{align}
and $A$ and $V$ are $G\times m$ matrices, with $m$ the minibatch size and $G$ the vector size after the projection $Y$ (see eq.~\eqref{eq:dilemma}), $z_j$ is the one-hot vector with $1$ at the $j$-th position and the index $n$ indicates the class token within the minibatch.

When we combine both the {\methodname} and the contrastive losses into a single cost we obtain
\begin{align}
    {\cal L}_\text{UNION} = \lambda_\text{\methodname}{\cal L}_\text{\methodname} + {\cal L}_\text{CNT},
\end{align}
which we minimize and where $\lambda_\text{\methodname}>0$ is a hyper parameter which we always set to $0.4$ .

\subsection{Implementation}
\noindent\textbf{Architecture.} We use Vision Transformers (ViT)~\cite{dosovitskiy2020image} with a patch size of $16\times 16$ pixels and an input image size of $224\times 224$ pixels, which gives a total of $(224/16)^2 = 196$ tokens. Due to computational limitations, we mostly use the small variant of the Vision Transformer (ViT-S) which has $12$ transformer blocks and $384$ channels. For the three baselines: 1) For MoCoV3~\cite{chen2021empirical} experiments, we use $12$ attention heads in each attention layer as specified in the official implementation. This is different from most ViT-S implementations, which use 6 heads. This does not change the total number of parameters of the model, but incurs a speed penalty. We use a 3-layer MLP for the projection and prediction heads with synchronized batch normalization. We also freeze the weights of the patch embedding layer for better stability; 2) SimCLR~\cite{chen2020simple} experiments are also conducted with the exact same settings, but without a teacher network and instead both augmentations are sparisified, misplaced and then fed to the student network; 3) For DINO~\cite{caron2021emerging} we used the official implementation and, whenever multi-crop is used, we have disabled random sparsity and used constant sparsity for the large crops and no sparsity for the small crops ($96\times 96$ images).\\

\noindent\textbf{Pre-training Setup.} For our main model, we pre-train {\methodname} on ImageNet-1K~\cite{deng2009imagenet} with the exact same hyper-parameters of MoCoV3 using three GeForce RTX 3090 GPUs for 100 epochs with a base batch size of 345. We set  $\lambda_\text{\methodname}$ to 0.4 and the probability of positional embedding mismatch $\theta=0.2$. We use sparsity ratios of 0\%, 40\%, 55\%, 65\% with $1\times$, $2\times$, $3\times$, $4\times$ base batch size and disable the {\methodname} loss when the input is dense.

To show the compatibility of the proposed method with other SSL methods, we also added two short runs for SimCLR and DINO with multi-cropping. For the DINO experiments we used ViT-Base to show that {\methodname} scales to larger models.
Since input sparsity allows for faster training, we also report results of {\methodname} variants with equal training time as the baselines.  
\\

\noindent\textbf{Linear Probing.} To evaluate the pre-trained features for image classification, we train a simple linear layer on top of \underline{frozen features}, without any data augmentation (Linear$_{F}$). Note that it is different from the standard linear probing, and we opt to use this method for its simplicity and speed. It is also more aligned with the end goal of representation learning. In all the linear probing experiments, we use the embedding of the CLS token of the last layer and perform a coarse grid search over learning rates, batch sizes and whether to normalize the data before feeding it to the linear layer or not (similarly to the added BatchNorm layer~\cite{ioffe2015batch} in MAE~\cite{he2021masked}). In contrast, DINO~\cite{caron2021emerging}, obtains its representation by concatenating the CLS token of the last four attention layers of the network.

\begin{table}[t]
    \centering
    \scalebox{0.84}{
        \begin{tabular}{@{}l@{\hspace{.3em}}cccccc@{}}
            \toprule
            Method & Epochs & Time & BS & {$k$-NN} & Linear$_{F}$ & Linear\\ \midrule
            SimCLR & 30 & 15.7h & 512 & 41.46 & 50.21 & -\\
            +Sparsity & 30 & 12.2h & 512 & 41.11 & 49.73 & -\\
            +\methodname & 30 & 12.2h & 512 & \bf 41.90 & \bf 50.71 & -\\
            \midrule
            DINO$^{\star\dagger}$ & 45 & 120.9h & 192 & 61.35 & 65.46 & -\\
            +Sparsity$^{\star\dagger}$ & 45 & 90.7h & 192 & 62.33 & 68.49 & -\\
            +\methodname$^{\star\dagger}$ & 45 & 90.7h & 192 & 62.48 & 68.55 & -\\
            +\methodname$^{\star\dagger\uparrow}$ & 60 & 121.0h & 192 & \bf 63.74 & \bf 69.43 & -\\ 
            \midrule
            MoCoV3 & 100 & 102.8h & 345 & 59.68 & 63.62 & 65.1\\
            +Sparsity & 100 & 68.4h & 345 & 61.64 & 65.16 & -\\
            +\methodname & 100 & 68.4h & 345 & 61.97 & 65.62 & 66.6\\
            +Sparsity$^{\uparrow}$ & 150 & 102.6h & 345 & 63.27 & 67.07 & -\\
            +\methodname$^{\uparrow}$ & 150 & 102.6h & 345 & \bf 64.69 & \bf 68.03 & -\\
            \midrule
            \midrule
            MoCoV3 & 300 & - & 4096 & 67.90 & 72.72 & 73.2\\
            DINO & 300 & - & 1024 & 67.9\phantom{0} & - & 72.5\\
            DINO$^{\dagger}$ & 800 & - & 1024 & 74.30 & 75.74 & 77.0\\
            Supervised & 300 & - & 1024 & - & - & 79.8\\
            \bottomrule
        \end{tabular}
    }
    \caption{ImageNet-1K Classification Transfer. The evaluation uses $k$-NN and linear probing with a ViT-S/16 or, where indicated, a ViT-Base/16 architecture. The $^\uparrow$ models are trained for a number of epochs, such that the total training time (see column Time) is the same as for the baseline methods. BS stands for Batch Size. $^\dagger$~models are trained with multi-crop. $^\star$ indicates ViT-Base/16 models}
    \label{tab:imagenet_fast}
\end{table}

\section{Experiments}

We evaluate the use of {\methodname} on several datasets, compare it to state-of-the-art (SotA) SSL baselines, and perform ablations to show the role of each loss component.
In each table, where we compare to an SSL baseline, we indicate the  baseline with a method name (\emph{e.g.}, MoCoV3~\cite{chen2021empirical}) and use a $\boldsymbol{+\{\text{\bf {\methodname}/sparsity}\}}$ to indicate that the baseline immediately above is combined with just sparsity or with the {\methodname} loss, which includes sparsity. We compare these two cases to show the added benefit of the {\methodname} positional classification loss over the lone sparsity.

\subsection{Classification on ImageNet-1K}

We show that {\methodname} leads to better representations for ImageNet-1K than prior SotA methods. Since this dataset has been used as a reference in SSL, it allows an easy comparison with previous work. 
In all tested cases, {\methodname} shows a consistent and significant improvement over the baseline it has been integrated with.
Notice that the improvement due to the positional loss, relative to the use of sparsity, becomes more significant with a longer training.\\

\begin{table*}[t]
    \centering
    \resizebox{1.0\textwidth}{!}{
        \begin{tabular}{@{}l@{\extracolsep{\fill}}ccccccccccccc|c@{}}
            \toprule
            & Aircraft & Caltech$_{101}$ & Cars & CIFAR$_{10}$ & CIFAR$_{100}$ & DTD & Flowers$_{102}$ & Food$_{101}$ & INat$_{19}$ & Pets & STL$_{10}$ & SVHN & Yoga$_{82}$ & Avg. \\
            \midrule
MoCoV3 & 38.70 & 87.35 & 28.72 & 91.97 & 75.09 & 64.63 & 91.67 & 67.97 & 33.30 & 84.63 & 95.76 & 64.56 & 56.41 & 67.75\\
-Position & 16.29 & 60.79 & 5.88 & 31.36 & 13.96 & 50.59 & 60.51 & 31.64 & 18.12 & 53.34 & 70.61 & 20.02 & 15.25 & 34.49\\
+Sparsity & 43.29 & 89.25 & 40.14 & 92.28 & 77.30 & 65.05 & 93.25 & 71.39 & 42.38 & 86.35 & 96.09 & 64.60 & 62.29 & 71.05\\
+\methodname & 44.43 & 89.55 & 42.32 & 93.03 & 77.56 & 64.47 & 93.41 & 71.94 & 43.76 & 85.88 & 96.08 & 65.41 & 63.76 & 71.66\\
+Sparsity$^{\uparrow}$ & 44.64 & 89.89 & 40.21 & 93.53 & 78.88 & \bf 65.48 & 94.21 & 72.96 & 42.84 & \bf 88.93 & \bf 96.90 & 64.52 & 63.24 & 72.02\\
+\methodname$^{\uparrow}$ & \bf 46.02 & \bf 90.29 & \bf 43.44 & \bf 94.20 & \bf 80.05 & 65.37 & \bf 94.47 & \bf 74.10 & \bf 44.13 & 88.53 & 96.75 & \bf 66.30 & \bf 64.90 & \bf 72.97\\
            \midrule
DINO & 45.66 & 88.30 & 47.07 & 90.61 & 74.06 & 66.22 & 94.54 & 73.00 & 47.36 & 85.83 & 96.62 & 53.74 & 59.90 & 70.99\\
+Sparsity & 46.83 & 89.16 & 48.45 & 92.47 & 77.00 & \bf 68.40 & 94.70 & 74.97 & 51.32 & 85.45 & 96.85 & 66.65 & 61.90 & 73.40\\
+\methodname & 46.86 & 89.58 & 49.04 & 92.46 & 77.22 & 67.50 & 95.20 & 74.90 & 52.51 & 85.91 & 97.06 & 70.01 & 62.87 & 73.93\\
+\methodname$^{\uparrow}$ & \bf 48.60 & \bf 89.66 & \bf 50.85 & \bf 93.39 & \bf 78.85 & 68.30 & \bf 95.38 & \bf 75.48 & \bf 52.71 & \bf 86.54 & \bf 97.41 & \bf 71.47 & \bf 64.00 & \bf 74.82\\
            \bottomrule
        \end{tabular}
    }
    \caption{Transfer learning for image classification on $13$ datasets. The $^\uparrow$ variants are trained for the same duration as the corresponding (non-sparse) baselines. $\boldsymbol{-}\text{\bf Position}$ is the case of MoCoV3, where the input tokens do not have the corresponding positional embeddings}
    \label{tab:many_shot_evaluation}
\end{table*}

\noindent\textbf{k-NN and Linear Probing.}
In Table~\ref{tab:imagenet_fast}, we evaluate the quality of the ImageNet-1K pre-trained features. We either use a weighted $k$ nearest neighbor classifier (we always use $k=20$)~\cite{Wu2018UnsupervisedFL} or a simple linear layer on top of a frozen backbone and \underline{frozen features}.
Since the use of sparsity has the added benefit of reducing the computational load at each iteration, we also show the actual training time. For example, with a ViT-Base/16 model and multi-crop, DINO $+$ {\methodname} (denoted with the $^\uparrow$ symbol) trains for 60 epochs in about the same time DINO trains for 45 epochs. This gives a significant advantage in performance. Furthermore, {\methodname} outperforms the baseline methods even if trained for the same number of epochs. The improvement under the same number of epochs is about $1-2$\% due to sparsity and $0.15-0.33$\% due to the positional classification loss for the k-NN evaluation. Similarly, it is about $2-3$\% due to sparsity and $0.06-0.46 $\% due to the positional classification loss for our linear probing. Notice that the boost due to the positional classification loss becomes more significant with more epochs (\emph{e.g}, for MoCoV3 and under the same running time, the k-NN evaluation shows a boost of $3.59$\% due to sparsity and an additional $1.42$\% due to the positional classification).

For the sake of completeness, we have also included best reported numbers for ViT-S/16 with significantly larger batch sizes and more training epochs.\\
\begin{table}[t]
    \centering
    \small
        \begin{tabular*}{\linewidth}{@{}c@{\hspace{.5em}}l@{\extracolsep{\fill}}cccc@{}}
        \toprule
        {} & {} & \multicolumn{2}{c}{ImageNet-1\%} & \multicolumn{2}{c}{ImageNet-10\%}\\
        {} & Method & {$k$-NN} & Linear$_{F}$ & {$k$-NN} & Linear$_{F}$\\ \midrule
        \multirow{6}{*}{\rot{Single-Crop}} & DINO & 40.60 & 45.24 & 52.95 & 58.35\\
        {} & MoCoV3 & 38.48 & 43.69 & 50.83 & 56.08\\
        {} & +Sparsity & 40.02 & 45.44 & 52.56 & 59.06\\
        {} & +\methodname & 41.64 & 47.95 & 53.15 & 60.00\\
        {} & +Sparsity$^{\uparrow}$ & 42.42 & 48.34 & 54.62 & 61.29\\
        {} & +\methodname$^{\uparrow}$ & \bf 45.62 & \bf 51.58 & \bf 56.66 & \bf 62.61\\
        \midrule
        \multirow{4}{*}{\rot{Multi-Crop}} & DINO & 41.79 & 46.88 & 53.00 & 59.48\\
        {} & +Sparsity & 42.36 & 48.65 & 53.61 & 62.33\\
        {} & +\methodname & 42.73 & 48.81 & 53.81 & 62.32\\
        {} & +\methodname$^{\uparrow}$ & \bf 43.87 & \bf 50.45 & \bf 55.29 & \bf 63.36\\
        \bottomrule
        \end{tabular*}
    \caption{Low-shot learning on ImageNet-1K.
    The $^\uparrow$ variants are trained for the same duration as the corresponding (non-sparse) baselines. In the Single-Crop case, DINO is shown only as a reference}
    \label{tab:imagenet_semisupervised}
\end{table}

\begin{table}[t]
    \centering
    \small
    \centering
    \setlength{\tabcolsep}{4.0pt}
        \begin{tabular}{@{}lcccccc@{}}
        \toprule  
        \multirow{2}{*}{Method} & \multicolumn{3}{c}{Seg. w/ Lin.} & \multicolumn{3}{c}{Seg. w/ UPerNet} \\
        \cmidrule(lr){2-4}\cmidrule(lr){5-7}
    & mIoU & mAcc & aAcc & mIoU & mAcc & aAcc \\
    \toprule
MoCoV3 & 12.44 & 15.91 & 65.95 & 32.13 & 43.37 & 76.79 \\
+Sparsity & 15.77 & 19.87 & 67.70 & 33.66 & 45.27 & 77.44 \\
+\methodname & 16.81 & 21.05 & 67.84 & 33.79 & 45.33 & 77.68 \\
+Sparsity$^{\uparrow}$ & 15.93 & 20.03 & 67.87 & 34.03 & 45.90 & 77.47 \\
+\methodname$^{\uparrow}$ & \bf 17.11 & \bf 21.48 & \bf 67.98 & \bf 34.98 & \bf 46.73 & \bf 77.97 \\
\midrule
DINO & 23.51 & 30.42 & 68.73 & 30.64 & 43.90 & 74.52 \\
+Sparsity & 26.75 & 34.20 & 71.61 & 33.82 & 47.01 & 76.56 \\
+\methodname & 27.78 & 35.63 & 72.63 & 34.11 & 47.50 & 76.73 \\
+\methodname$^{\uparrow}$ & \bf 28.72 & \bf 36.72 & \bf 72.79 & \bf 34.87 & \bf 47.95 & \bf 77.61 \\
        \bottomrule
        \end{tabular}
    \caption{Semantic Segmentation on ADE20K. The $^\uparrow$ variants are trained for the same duration as the corresponding (non-sparse) baselines}
    \label{tab:semantic_segmentation}
\end{table}

\begin{table}[t]
    \centering
    \small
    \begin{tabular*}{\linewidth}{@{}l@{\extracolsep{\fill}}cccc@{}}
    \toprule
    {} & \multicolumn{3}{c}{DAVIS} & VOC12 \\
    Method & $ (\mathcal{J}$\&$\mathcal{F})_m$ & $\mathcal{J}_m$ & $\mathcal{F}_m$ & Jac.$_{sim.}$\\
    \midrule
    MoCoV3 & 58.28 & 57.46 & 59.09 & 46.50 \\
    +Sparsity & 58.94 & 57.05 & 60.83 & 45.34 \\
    +\methodname & \bf 60.00 & \bf 57.99 & \bf 62.02 & 48.89 \\
    +Sparsity$^{\uparrow}$ & 58.03 & 56.74 & 59.33 & 45.93 \\
    +\methodname$^{\uparrow}$ & 59.84 & 57.98 & 61.69 & \bf 49.36 \\
    \midrule
    DINO & 57.01 & 55.13 & 58.90 & 41.60 \\
    +Sparsity & 56.83 & 54.84 & 58.81 & 40.03 \\
    +\methodname & \bf 57.60 & \bf 55.39 & \bf 59.80 & 39.71 \\
    +\methodname$^{\uparrow}$ & 57.25 & 55.18 & 59.31 & \bf 44.14 \\
    \bottomrule
    \end{tabular*}
    \caption{Unsupervised object segmentation. We show the mean region similarity $\mathcal{J}_m$ and the mean contour-based accuracy $\mathcal{F}_m$ for DAVIS, and the Jaccard similarity for VOC12. The $^\uparrow$ variants are trained for the same duration as the corresponding (non-sparse) baselines}
    \label{tab:video}
\end{table}

\noindent\textbf{Low-shot learning.}
In Table~\ref{tab:imagenet_semisupervised}, we simulate transfers to small datasets. With reference to ImageNet, we use the model pre-trained on the whole unlabeled dataset, train a linear layer on top of the frozen features of the 1\% or 10\% subsets \cite{chen2020simple} and then evaluate the results on the whole validation set. The results show that adding {\methodname} to MoCoV3 or DINO yields a more label-efficient representation than with the corresponding baselines. Notice that in this implementation {\methodname} is based on MoCoV3, which, as was observed in DINO~\cite{caron2021emerging}, has a consistently worse $k$-NN accuracy than DINO. Nonetheless, the addition of {\methodname} can more than compensate for the performance gap.

\subsection{Downstream Tasks}

We evaluate {\methodname} on several datasets to assess its generalization capability across different classification and detection tasks. While {\methodname} improves the performance over the baselines in all the datasets, the most significant improvement seems to occur for more shape-based tasks, such as pose classification. The evaluation on object segmentation, which is a dense downstream task, illustrates the representation captured by the non-CLS tokens.\\

\noindent\textbf{Transfer Learning.}
In Table~\ref{tab:many_shot_evaluation}, we evaluate the transfer capability of our representations for image classification on several datasets. We use: Aircraft \cite{maji13fine-grained}, Caltech$_{101}$ \cite{FeiFei2004LearningGV}, Cars \cite{KrauseStarkDengFei-Fei_3DRR2013}, CIFAR$_{10}$ \cite{Krizhevsky2009LearningML}, CIFAR$_{100}$ \cite{Krizhevsky2009LearningML}, DTD \cite{cimpoi14describing}, Flowers$_{102}$ \cite{Nilsback2008AutomatedFC}, Food$_{101}$ \cite{bossard14}, INat$_{19}$ \cite{inaturalist19}, Pets \cite{parkhi12a}, STL$_{10}$ \cite{Coates2011AnAO}, SVHN \cite{Netzer2011}, and Yoga$_{82}$ \cite{verma2020yoga}. We train a linear layer on top of the frozen features to accelerate the process. {\methodname} performs well in transfer learning across all datasets and significantly more on datasets with  shape-based tasks, such as Yoga$_{82}$~\cite{verma2020yoga} (for yoga position classification).

We also try to measure approximately how much shape matters in each dataset. 
We evaluate MoCoV3 with tokens without their position embedding. For simplicity, we use the same pre-trained MoCoV3 used throughout the experiments (although one should ideally use a MoCoV3 trained without position embeddings).
We indicate this case with $\boldsymbol{-\text{\bf Position}}$ in Table~\ref{tab:many_shot_evaluation}. Without position embedding these features are equivalent to a bag of features. We can see that the improvement due to {\methodname} relative to the baseline MoCoV3 follows the corresponding relative degradation due to the bag of features representation. This suggests that {\methodname} tends to generalize better on datasets with shape-based tasks.\\

\noindent\textbf{Semantic Segmentation on ADE20K.}
In Table~\ref{tab:semantic_segmentation}, we show the evaluation of {\methodname} on semantic segmentation.
This is a task that strongly relates to the shape of objects. Thus, we expect to see a significant improvement from a boost in the shape discriminability. The semantic segmentation capability of self-supervised methods is usually evaluated by fine-tuning the model with an extra decoder. For that we use UPerNet~\cite{xiao2018unified} on the ADE20K~\cite{zhou2017scene} dataset and train the model for $160$K iterations with a batch size of 2 for ViT-Base and 8 for ViT-Small. We also follow the evaluation protocol of iBOT~\cite{zhou2021ibot} and just train a linear layer (for $160$K iterations and a batch size of 16) for semantic segmentation with a frozen backend to directly assess the per-token representation. The results show that {\methodname} is also better than the baseline models for dense classification tasks. It yields remarkable mIoU improvements of $4.6$\% against MoCoV3 and of $5.2$\% against DINO in the linear settings and under the same training time.\\

\noindent\textbf{Unsupervised Object Segmentation.}
In Table~\ref{tab:video}, we evaluate the single frame object segmentation task. We use the mask generated from the attention of the CLS token (thresholded to keep 0.9 of the mass) as in DINO~\cite{caron2021emerging}, and report the Jaccard similarity between the ground truth and the mask evaluated on the validation set of PASCAL-VOC12~\cite{Everingham2009ThePV}. For the videos we use the DAVIS-2017 video instance segmentation benchmark~\cite{Pont-Tuset_arXiv_2017} and by following the protocol introduced in Space-time by Jabri et al.~\cite{jabri2020space} we segment scenes via the nearest neighbor propagation of the mask. In these evaluations, the role of the positional classification loss seems to be more important than sparsity alone.\\

\noindent\textbf{Humanoid Vision Engine Benchmark.} We also use the newly introduced HVE~\cite{ge2022contributions} to evaluate our shape bias in Table~\ref{tab:hve}. In HVE Shape dataset, the input images are only the depth map of the foreground object which only contains shape information. We see that {\methodname} outperforms the base model which confirms our hypothesis that {\methodname} can focus on shape. For the HVE Texture, only four grey scaled random crops of the foreground object are concatenated and fed as input, so predicting the right class requires high texture discriminability. Results on HVE Texture show that {\methodname}'s better shape understanding did not harm the texture discriminability. \\
\begin{table}[t]
    \centering
        \begin{tabular}{lcc}
    \toprule
    Method & Shape Accuracy & Texture Accuracy\\
    \midrule
MoCoV3 & 80.78 & 82.66\\
+Sparsity & 82.55 & 81.78\\
+\methodname & \bf 83.58 & 82.11\\
+Sparsity$^{\uparrow}$ & 82.72 & 82.77\\
+\methodname$^{\uparrow}$ & 83.52 & \bf 83.82\\
\midrule
DINO & 80.84 & 79.47\\
+Sparsity & 83.18 & 81.01\\
+\methodname & 83.58 & 80.79\\
+\methodname$^{\uparrow}$ & \bf 83.64 & \bf 81.45\\
    \bottomrule
    \end{tabular}
    \caption{Humanoid Vision Engine Benchmark~\cite{ge2022contributions} results. The $^\uparrow$ models are trained for a number of epochs, such that the total training time is the same as for the baseline methods}
    \label{tab:hve}
\end{table}

\noindent\textbf{Robustness against Background Change.} Following the background challenge evaluation metric~\cite{xiao2020noise}, we compute the classification accuracy of the model on a subset of ImageNet (IN-9) by changing the background and foreground. As shown in Table~\ref{tab:background}, in O/N.F. (Only/No Foreground), M.S/R/N. (Mixed Same/Random/Next), where the foreground is visible or accurately masked out, we outperform the base model. When the foreground is not visible (O.BB. (Only Background with foreground box Blacked out) and O.BT. (Only Background with foreground replaced with Tiled background)) the model performs correctly and does not just rely on the background for image classification.

\begin{table*}[t]
\centering
\begin{tabular}{@{}lcccccccc@{}}
\multirow{2}{*}{Method} & \multicolumn{7}{c}{Background Change} & Clean \\
\cmidrule(lr){2-8}\cmidrule(lr){9-9}
& \it M.N.$(\uparrow)$ & \it M.R.$(\uparrow)$ & \it M.S.$(\uparrow)$ & \it N.F.$(\uparrow)$ & \it O.BB.$(\downarrow)$ & \it O.BT.$(\downarrow)$ & \it O.F.$(\uparrow)$ & IN-9$(\uparrow)$\\
\toprule
MoCoV3 & 64.52 & 65.68 & 78.57 & 38.69 & 9.41 & 10.67 & 77.80 & 91.65\\
+Sparsity & 65.53 & 67.75 & 80.25 & 38.72 & 9.48 & 10.40 & 78.15 & 92.52\\
+\methodname & 65.19 & 68.37 & 79.63 & 39.19 & \bf 8.42 & \bf 9.68 & 78.37 & 92.00\\
+Sparsity$^{\uparrow}$ & 66.25 & 69.60 & 81.26 & 40.25 & 10.99 & 10.25 & 80.10 & 92.77\\
+\methodname$^{\uparrow}$ & \bf 68.86 & \bf 71.16 & \bf 81.85 & \bf 40.40 & 8.69 & 10.64 & \bf 82.42 & \bf 93.46\\
\midrule
DINO & 65.56 & 68.94 & 79.95 & 33.28 & 9.70 & \bf 9.90 & 80.99 & 92.17\\
+Sparsity & 67.68 & 71.28 & 82.10 & 35.23 & 8.89 & 11.11 & 83.26 & 93.43\\
+\methodname & \bf 69.58 & 71.85 & 82.89 & 36.02 & 9.31 & 10.47 & 83.75 & 93.06\\
+\methodname$^{\uparrow}$ & 69.38 & \bf 73.75 & \bf 82.94 & \bf 38.54 & \bf 8.81 & \bf 9.90 & \bf 84.69 & \bf 93.93\\
\bottomrule
\end{tabular}
\caption{Robustness of pre-trained models against background changes. The $^\uparrow$ models are trained for a number of epochs, such that the total training time is the same as for the baseline methods}
\label{tab:background}
\end{table*}

\begin{table}[t]
\centering
    \small
\begin{tabular*}{\linewidth}{@{}l@{\extracolsep{\fill}}cr@{}}
\toprule
Sampling Method & $k$-NN & Linear\\
\hline
Importance Based & 71.88 & 76.76\\
Random & \bf 73.98 & \bf 77.78\\
\bottomrule
\end{tabular*}
\caption{Token dropping policy. Results are evaluated on IN100}
\label{tab:drop_policy}
\end{table}

\begin{table}[t]
\centering
    \small
\begin{tabular*}{\linewidth}{@{}l@{\extracolsep{\fill}}cccc@{}}
\toprule
{} & \multicolumn{2}{c}{IN100} & \multicolumn{2}{c}{IN-1K}\\
Sparsity & $k$-NN & Linear & $k$-NN & Linear\\
\hline
0\% (Dense) & \bf 76.16 & 77.50 & 53.27 & 58.20\\
75\% & 73.98 & 77.78 & 52.99 & 57.90\\
Random & 74.46 & \bf 78.82 & \bf 55.71 & \bf 59.55\\
\bottomrule
\end{tabular*}
\caption{Random Dropping Ratio. Results on the left are evaluated on IN100 and on the right on IN-1K}
\label{tab:random_drop_ratio}
\end{table}

\begin{table}[t]
\centering
\small
\begin{tabular*}{\linewidth}{@{}l@{\extracolsep{\fill}}cc|cc|c@{}}
\toprule
{} & \multicolumn{2}{c}{IN-1K} & \multicolumn{2}{c}{Yoga82} & {}\\
{} & $k$-NN & Linear & $k$-NN & Linear & MD Acc.\\
\hline
MoCoV3 & 53.27 & 58.20 & 31.60 & 51.27 & -\\
+MD & 54.18 & 58.78 & 35.78 & 54.53 & 100.00\\
+Sparsity & \bf 55.71 & 59.55 & 32.73 & 50.90 & -\\
+Both & 55.63 & \bf 59.84 & \bf 35.94 & \bf 57.26 & 96.21\\
\bottomrule
\end{tabular*}
\caption{Mismatch Detection (MD). Detecting misplaced tokens for dense inputs is easily solved, but still improves the model's performance on shape based tasks. Note that adding both MD and Sparsity to the base model is the same as \methodname}
\label{tab:electra_helps}
\end{table}

\subsection{Ablations}

In these experiments, we want to validate empirically a number of choices: 1) we ask how much the trained model is robust to occlusions (sparsity) and positional errors; 2) whether the selection of tokens should be random or guided; 3) whether the ratio of dropped tokens should remain constant in time or instead vary; 4) what the relevance of the positional classification loss is; 5) the impact of the number of positional errors used during training; 6) whether other design variations are more effective than {\methodname}.

Ablation studies are conducted either on ImageNet100 (IN100) or ImageNet-1K (IN-1K). For the smaller dataset we train the dense models for $300$ epochs and the sparse models for $450$ epochs (with the same hardware and time settings). For IN-1K experiments we train all models for $50$ epochs with MoCoV3 unless stated otherwise.\\

\noindent\textbf{Token Dropping Policy.}
In Table~\ref{tab:drop_policy}, we compare the case of dropping the tokens that are less important based on the attention of the teacher network~\cite{li2021mst} compared to randomly dropping the tokens. Results show that simple random dropping works well and there is no need to introduce extra complexity to the policy.\\

\begin{table}[t]
\centering
    \small
\begin{tabular*}{\linewidth}{@{}l@{\extracolsep{\fill}}cccc@{}}
\toprule
{} & \multicolumn{2}{c}{IN-1K} & \multicolumn{2}{c}{Yoga82}\\
$\theta$ & $k$-NN & Linear & $k$-NN & Linear\\
\hline
0.3 & 55.34 & 59.79 & 34.95 & 56.63\\
0.2 & \bf 55.63 & \bf 59.84 & \bf 35.94 & \bf 57.26\\
\bottomrule
\end{tabular*}
\caption{Mismatch Probability. Too much mismatch hurts performance}
\label{tab:mismatch_prob}
\end{table}

\begin{table}[t!]
\centering
    \small
\begin{tabular*}{\linewidth}{@{}l@{\extracolsep{\fill}}cc|cc@{}}
\toprule
{} & \multicolumn{2}{c}{IN-1K} & \multicolumn{2}{c}{Yoga82}\\
Task & $k$-NN & Linear & $k$-NN & Linear\\
\hline
None (MoCoV3) & 53.27 & 58.20 & 31.60 & 51.27\\
Pos. Correction & 54.77 & 58.95 & 35.74 & 56.15\\
Partial Jigsaw & \bf 55.72 & 59.19 & 34.77 & 56.79\\
Flip Detection & 55.69 & 59.59 & 35.09 & 55.00\\
\methodname & 55.63 & \bf 59.84 & \bf 35.94 & \bf 57.26\\
\bottomrule
\end{tabular*}
\caption{Variants of the loss. Although the variants improve the performance wrt the baseline, {\methodname} is the most effective one}
\label{tab:task_variants}
\end{table}

\noindent\textbf{Randomized Dropping Ratio.}
In Table~\ref{tab:random_drop_ratio}, we verify that a randomized dropping ratio is better than a constant one. We conducted two experiments: one on IN100 and one on IN-1K. The results show that a randomized dropping ratio performs better than a constant one. On the more difficult IN-1K dataset, just applying sparsity is worse than using the dense model. Only with a random dropping ratio can the sparse model outperform the dense model.\\

\noindent\textbf{Position Classification Loss.} In Table~\ref{tab:electra_helps}, we verify that the position classification loss helps, by training a dense model with position mismatch detection. Surprisingly, even though the Mismatch Detection (MD) (\emph{i.e.}, the average classification accuracy of the token locations -- see ``MD Acc.'' in Table~\ref{tab:electra_helps}) is easily solved (it achieves $100$\% in the dense case),
the dense model can still improve the performance of the model on a downstream task. The performance improvement for a task like in $\text{Yoga}_{82}$, which requires a better understanding of shape, is quite significant both with the dense and randomized sparsity inputs.\\

\noindent\textbf{Mismatch Probability.} The probability of a positional embedding mismatch $\theta$ is one of the hyper-parameters of {\methodname}. Early in our experiments, we found out that $20$\% is much better than $15$\% (which is used by Electra~\cite{clark2020electra}), probably due to the higher information redundancy in images compared to text. In Table~\ref{tab:mismatch_prob}, we show that $\theta=30$\% yields worse performance than the default $\theta=20$\%.\\

\noindent\textbf{{\methodname} Variants.}
We also tried some variants of \methodname. Instead of just detecting the misplaced tokens, we predict the right position (as a classification task of $196$ classes). The other variant, which we call \emph{Partial Jigsaw}, is to feed some tokens without position encoding and ask the network to predict their position given the other (sparse) correctly position-encoded tokens. Lastly, instead of corrupting the position, one can corrupt the content of a patch. Instead of using complex methods like inpainting we simply horizontally flip some of the patches and use the binary cross-entropy as our loss. Table~\ref{tab:task_variants} shows that even though all of these methods do help in terms of shape discrimination, {\methodname} is the one with the best performance both on IN-1K and $\text{Yoga}_{82}$.\\

\noindent\textbf{Timing.}
To show the efficiency of the proposed method, we ran SimCLR, MoCoV3, DINO with and without multi-crop on 4 GPUs and reported the epoch times in table~\ref{tab:timing}.\\

\begin{table}[t]
    \centering
    \centering
    \begin{adjustbox}{width=\columnwidth}
    \begin{tabular}{lccc}
    \toprule
    Method & BatchSize & EpochTime & MaxMem(GB)\\
    \midrule
    SimCLR & 640 & 21:35 & 23.65\\
    +\methodname & 1680 & 18:20 ($\times$0.85) & 23.17\\
    MoCoV3 & 656 & 49:08 & 23.41\\
    +\methodname & 1664 & 32:11 ($\times$0.65) & 23.55\\
    DINO & 576 & 37:57 & 22.73\\
    +\methodname & 1184 & 24:13 ($\times$0.64) & 22.79\\
    DINO(MC)$^{\dagger}$ & 144 & 3:07:21 & 22.85\\
    +\methodname$^{\dagger}$ & 216 & 2:15:52 ($\times$0.72) & 23.69\\
    \bottomrule
    \end{tabular}
    \end{adjustbox}
    \caption{Timing and memory usage of training ViT-Small models with four RTX Geforce 3090 GPUs. MC stands for Multi-Crop. $^{\dagger}$ models use ViT-Base}
    \label{tab:timing}
\end{table}

\noindent\textbf{Combining with MAE.}
To show the general applicability of our proposed method to masked models, we misplaced some of the MAE~\cite{he2021masked} inputs and added {\methodname} loss to the \emph{encoder} of MAE in addition to the reconstruction loss of the decoder. Both MAE and {\methodname} are trained for 200 epochs on IN-100 (using the exact same hyperparameters of the official repository) and  results in table~\ref{tab:mae} show that we can outperform MAE both in terms of linear probe and finetuning.\\

\begin{table}[t]
\centering
    \small
\begin{tabular*}{\linewidth}{@{}l@{\extracolsep{\fill}}cr@{}}
\toprule
Method & Linear & Finetune\\
\hline
MAE & 37.30 & 82.60\\
+\methodname & \bf 39.06 & \bf 83.30\\
\bottomrule
\end{tabular*}
\caption{Combining {\methodname} with MAE. Results are evaluated on IN100 after pretraining for 200 epochs and using ViT-Base}
\label{tab:mae}
\end{table}

\noindent\textbf{Longer Pretraining.}
We pretrain MoCoV3 and {\methodname} for 1000 epochs on IN-100 and evaluate their linear performance to see whether the benefits of {\methodname} still hold for longer pretrainings. Results in table~\ref{tab:longer} show that indeed {\methodname} always performs better than the baseline even with longer pretraining.\\

\begin{table}[t]
\centering
    \small
\begin{tabular*}{\linewidth}{@{}l@{\extracolsep{\fill}}cr@{}}
\toprule
Method & 300 Epochs & 1000 Epochs\\
\hline
MoCoV3 & 77.50 & 79.76\\
+\methodname & \bf78.82 & \bf81.26\\
\bottomrule
\end{tabular*}
\caption{Longer Pretraining on IN-100. Linear accuracies on IN100 after pretraining for 300 and 1000 epochs}
\label{tab:longer}
\end{table}

\noindent\textbf{Weaker Data Augmentations.}
One of the most important factors for the performance of contrastive learners is the data augmentation. In this short experiment (50 epochs of pretraining, and 70 epochs of linear training) we only used random resized cropping (like MAE~\cite{he2021masked}) on IN-1K for both MoCoV3 and {\methodname}. Linear probe accuracy of {\methodname} is \textbf{44.48\%} and for MoCoV3 it is \textbf{29.65\%} (Note that a 100 epoch pretrained ResNet-50~\cite{He2016DeepRL} with SimCLR~\cite{chen2020simple} gets 33.1\% accuracy). This huge gap shows that {\methodname} is a generic method for representation learning and does not completely depend on the contrastive component of the loss.

\section{Conclusions}
We introduced a novel SSL method based on a position classification pseudo-task and a contrastive loss. We showed that awareness of the relative location of tiles of the input image is important for generalization and in particular when fine-tuning on shape-based downstream tasks.
Since our method is based on the ViT architecture, we introduce sparsity in the input (\emph{i.e.}, dropping image tiles), to both speed up the training and also to avoid trivial degenerate learning.

\section{Acknowledgments}
This work was supported by grant 200020\_188690 of the Swiss National Science Foundation (SNSF) and an Adobe award.

\small
\bibliography{aaai23}

\begin{thebibliography}{68}
\providecommand{\natexlab}[1]{#1}

\bibitem[{Akbari et~al.(2021)Akbari, Yuan, Qian, Chuang, Chang, Cui, and
  Gong}]{akbari2021vatt}
Akbari, H.; Yuan, L.; Qian, R.; Chuang, W.-H.; Chang, S.-F.; Cui, Y.; and Gong,
  B. 2021.
\newblock Vatt: Transformers for multimodal self-supervised learning from raw
  video, audio and text.
\newblock \emph{Advances in Neural Information Processing Systems}, 34.

\bibitem[{Bao, Dong, and Wei(2021)}]{bao2021beit}
Bao, H.; Dong, L.; and Wei, F. 2021.
\newblock Beit: Bert pre-training of image transformers.
\newblock \emph{arXiv preprint arXiv:2106.08254}.

\bibitem[{Bossard, Guillaumin, and Van~Gool(2014)}]{bossard14}
Bossard, L.; Guillaumin, M.; and Van~Gool, L. 2014.
\newblock Food-101 -- Mining Discriminative Components with Random Forests.
\newblock In \emph{European Conference on Computer Vision}.

\bibitem[{Caron et~al.(2020)Caron, Misra, Mairal, Goyal, Bojanowski, and
  Joulin}]{caron2020unsupervised}
Caron, M.; Misra, I.; Mairal, J.; Goyal, P.; Bojanowski, P.; and Joulin, A.
  2020.
\newblock Unsupervised learning of visual features by contrasting cluster
  assignments.
\newblock \emph{Advances in Neural Information Processing Systems}, 33:
  9912--9924.

\bibitem[{Caron et~al.(2021)Caron, Touvron, Misra, J{\'e}gou, Mairal,
  Bojanowski, and Joulin}]{caron2021emerging}
Caron, M.; Touvron, H.; Misra, I.; J{\'e}gou, H.; Mairal, J.; Bojanowski, P.;
  and Joulin, A. 2021.
\newblock Emerging properties in self-supervised vision transformers.
\newblock In \emph{Proceedings of the IEEE/CVF International Conference on
  Computer Vision}, 9650--9660.

\bibitem[{Chen et~al.(2020{\natexlab{a}})Chen, Kornblith, Norouzi, and
  Hinton}]{chen2020simple}
Chen, T.; Kornblith, S.; Norouzi, M.; and Hinton, G. 2020{\natexlab{a}}.
\newblock A simple framework for contrastive learning of visual
  representations.
\newblock In \emph{International conference on machine learning}, 1597--1607.
  PMLR.

\bibitem[{Chen et~al.(2022)Chen, Ding, Wang, Xin, Mo, Wang, Han, Luo, Zeng, and
  Wang}]{chen2022context}
Chen, X.; Ding, M.; Wang, X.; Xin, Y.; Mo, S.; Wang, Y.; Han, S.; Luo, P.;
  Zeng, G.; and Wang, J. 2022.
\newblock Context Autoencoder for Self-Supervised Representation Learning.
\newblock \emph{arXiv preprint arXiv:2202.03026}.

\bibitem[{Chen et~al.(2020{\natexlab{b}})Chen, Fan, Girshick, and
  He}]{chen2020improved}
Chen, X.; Fan, H.; Girshick, R.; and He, K. 2020{\natexlab{b}}.
\newblock Improved baselines with momentum contrastive learning.
\newblock \emph{arXiv preprint arXiv:2003.04297}.

\bibitem[{Chen and He(2020)}]{chen2020exploring}
Chen, X.; and He, K. 2020.
\newblock Exploring Simple Siamese Representation Learning.
\newblock \emph{arXiv preprint arXiv:2011.10566}.

\bibitem[{Chen, Xie, and He(2021)}]{chen2021empirical}
Chen, X.; Xie, S.; and He, K. 2021.
\newblock An empirical study of training self-supervised vision transformers.
\newblock In \emph{Proceedings of the IEEE/CVF International Conference on
  Computer Vision}, 9640--9649.

\bibitem[{Cimpoi et~al.(2014)Cimpoi, Maji, Kokkinos, Mohamed, , and
  Vedaldi}]{cimpoi14describing}
Cimpoi, M.; Maji, S.; Kokkinos, I.; Mohamed, S.; ; and Vedaldi, A. 2014.
\newblock Describing Textures in the Wild.
\newblock In \emph{Proceedings of the {IEEE} Conf. on Computer Vision and
  Pattern Recognition ({CVPR})}.

\bibitem[{Clark et~al.(2020{\natexlab{a}})Clark, Luong, Le, and
  Manning}]{clark2020electra}
Clark, K.; Luong, M.-T.; Le, Q.~V.; and Manning, C.~D. 2020{\natexlab{a}}.
\newblock Electra: Pre-training text encoders as discriminators rather than
  generators.
\newblock \emph{arXiv preprint arXiv:2003.10555}.

\bibitem[{Clark et~al.(2020{\natexlab{b}})Clark, Luong, Le, and
  Manning}]{clark2020pre}
Clark, K.; Luong, M.-T.; Le, Q.~V.; and Manning, C.~D. 2020{\natexlab{b}}.
\newblock Pre-training transformers as energy-based cloze models.
\newblock \emph{arXiv preprint arXiv:2012.08561}.

\bibitem[{Coates, Ng, and Lee(2011)}]{Coates2011AnAO}
Coates, A.; Ng, A.; and Lee, H. 2011.
\newblock An Analysis of Single-Layer Networks in Unsupervised Feature
  Learning.
\newblock In \emph{AISTATS}.

\bibitem[{Deng et~al.(2009)Deng, Dong, Socher, Li, Li, and
  Fei-Fei}]{deng2009imagenet}
Deng, J.; Dong, W.; Socher, R.; Li, L.-J.; Li, K.; and Fei-Fei, L. 2009.
\newblock Imagenet: A large-scale hierarchical image database.
\newblock In \emph{2009 IEEE conference on computer vision and pattern
  recognition}, 248--255. Ieee.

\bibitem[{Devlin et~al.(2018)Devlin, Chang, Lee, and
  Toutanova}]{devlin2018bert}
Devlin, J.; Chang, M.-W.; Lee, K.; and Toutanova, K. 2018.
\newblock Bert: Pre-training of deep bidirectional transformers for language
  understanding.
\newblock \emph{arXiv preprint arXiv:1810.04805}.

\bibitem[{Doersch, Gupta, and Efros(2015)}]{doersch2015unsupervised}
Doersch, C.; Gupta, A.; and Efros, A.~A. 2015.
\newblock Unsupervised visual representation learning by context prediction.
\newblock In \emph{Proceedings of the IEEE international conference on computer
  vision}, 1422--1430.

\bibitem[{Dosovitskiy et~al.(2020)Dosovitskiy, Beyer, Kolesnikov, Weissenborn,
  Zhai, Unterthiner, Dehghani, Minderer, Heigold, Gelly
  et~al.}]{dosovitskiy2020image}
Dosovitskiy, A.; Beyer, L.; Kolesnikov, A.; Weissenborn, D.; Zhai, X.;
  Unterthiner, T.; Dehghani, M.; Minderer, M.; Heigold, G.; Gelly, S.; et~al.
  2020.
\newblock An image is worth 16x16 words: Transformers for image recognition at
  scale.
\newblock \emph{arXiv preprint arXiv:2010.11929}.

\bibitem[{Dosovitskiy et~al.(2015)Dosovitskiy, Fischer, Springenberg,
  Riedmiller, and Brox}]{dosovitskiy2015discriminative}
Dosovitskiy, A.; Fischer, P.; Springenberg, J.~T.; Riedmiller, M.; and Brox, T.
  2015.
\newblock Discriminative unsupervised feature learning with exemplar
  convolutional neural networks.
\newblock \emph{IEEE transactions on pattern analysis and machine
  intelligence}, 38(9): 1734--1747.

\bibitem[{Dwibedi et~al.(2021)Dwibedi, Aytar, Tompson, Sermanet, and
  Zisserman}]{dwibedi2021little}
Dwibedi, D.; Aytar, Y.; Tompson, J.; Sermanet, P.; and Zisserman, A. 2021.
\newblock With a little help from my friends: Nearest-neighbor contrastive
  learning of visual representations.
\newblock In \emph{Proceedings of the IEEE/CVF International Conference on
  Computer Vision}, 9588--9597.

\bibitem[{El-Nouby et~al.(2021)El-Nouby, Izacard, Touvron, Laptev, Jegou, and
  Grave}]{el2021large}
El-Nouby, A.; Izacard, G.; Touvron, H.; Laptev, I.; Jegou, H.; and Grave, E.
  2021.
\newblock Are Large-scale Datasets Necessary for Self-Supervised Pre-training?
\newblock \emph{arXiv preprint arXiv:2112.10740}.

\bibitem[{Everingham et~al.(2009)Everingham, Gool, Williams, Winn, and
  Zisserman}]{Everingham2009ThePV}
Everingham, M.; Gool, L.~V.; Williams, C. K.~I.; Winn, J.~M.; and Zisserman, A.
  2009.
\newblock The Pascal Visual Object Classes (VOC) Challenge.
\newblock \emph{International Journal of Computer Vision}, 88: 303--338.

\bibitem[{Fang et~al.(2022)Fang, Dong, Bao, Wang, and
  Wei}]{Fang2022CorruptedIM}
Fang, Y.; Dong, L.; Bao, H.; Wang, X.; and Wei, F. 2022.
\newblock Corrupted Image Modeling for Self-Supervised Visual Pre-Training.
\newblock \emph{ArXiv}, abs/2202.03382.

\bibitem[{Fei-Fei, Fergus, and Perona(2004)}]{FeiFei2004LearningGV}
Fei-Fei, L.; Fergus, R.; and Perona, P. 2004.
\newblock Learning Generative Visual Models from Few Training Examples: An
  Incremental Bayesian Approach Tested on 101 Object Categories.
\newblock \emph{2004 Conference on Computer Vision and Pattern Recognition
  Workshop}, 178--178.

\bibitem[{Ge et~al.(2022)Ge, Xiao, Xu, Wang, and Itti}]{ge2022contributions}
Ge, Y.; Xiao, Y.; Xu, Z.; Wang, X.; and Itti, L. 2022.
\newblock Contributions of Shape, Texture, and Color in Visual Recognition.
\newblock \emph{arXiv preprint arXiv:2207.09510}.

\bibitem[{Geirhos et~al.(2018)Geirhos, Rubisch, Michaelis, Bethge, Wichmann,
  and Brendel}]{geirhos2018imagenet}
Geirhos, R.; Rubisch, P.; Michaelis, C.; Bethge, M.; Wichmann, F.~A.; and
  Brendel, W. 2018.
\newblock ImageNet-trained CNNs are biased towards texture; increasing shape
  bias improves accuracy and robustness.
\newblock \emph{arXiv preprint arXiv:1811.12231}.

\bibitem[{Gidaris, Singh, and Komodakis(2018)}]{gidaris2018unsupervised}
Gidaris, S.; Singh, P.; and Komodakis, N. 2018.
\newblock Unsupervised representation learning by predicting image rotations.
\newblock \emph{arXiv preprint arXiv:1803.07728}.

\bibitem[{Girshick et~al.(2014)Girshick, Donahue, Darrell, and
  Malik}]{girshick2014rich}
Girshick, R.; Donahue, J.; Darrell, T.; and Malik, J. 2014.
\newblock Rich feature hierarchies for accurate object detection and semantic
  segmentation.
\newblock In \emph{Proceedings of the IEEE conference on computer vision and
  pattern recognition}, 580--587.

\bibitem[{Grill et~al.(2020)Grill, Strub, Altch'e, Tallec, Richemond,
  Buchatskaya, Doersch, Pires, Guo, Azar, Piot, Kavukcuoglu, Munos, and
  Valko}]{Grill2020BootstrapYO}
Grill, J.-B.; Strub, F.; Altch'e, F.; Tallec, C.; Richemond, P.~H.;
  Buchatskaya, E.; Doersch, C.; Pires, B.~{\'A}.; Guo, Z.~D.; Azar, M.~G.;
  Piot, B.; Kavukcuoglu, K.; Munos, R.; and Valko, M. 2020.
\newblock Bootstrap Your Own Latent: A New Approach to Self-Supervised
  Learning.
\newblock \emph{ArXiv}, abs/2006.07733.

\bibitem[{He et~al.(2021)He, Chen, Xie, Li, Doll{\'a}r, and
  Girshick}]{he2021masked}
He, K.; Chen, X.; Xie, S.; Li, Y.; Doll{\'a}r, P.; and Girshick, R. 2021.
\newblock Masked autoencoders are scalable vision learners.
\newblock \emph{arXiv preprint arXiv:2111.06377}.

\bibitem[{He et~al.(2020)He, Fan, Wu, Xie, and Girshick}]{he2020momentum}
He, K.; Fan, H.; Wu, Y.; Xie, S.; and Girshick, R. 2020.
\newblock Momentum contrast for unsupervised visual representation learning.
\newblock In \emph{Proceedings of the IEEE/CVF conference on computer vision
  and pattern recognition}, 9729--9738.

\bibitem[{He et~al.(2016)He, Zhang, Ren, and Sun}]{He2016DeepRL}
He, K.; Zhang, X.; Ren, S.; and Sun, J. 2016.
\newblock Deep Residual Learning for Image Recognition.
\newblock \emph{2016 IEEE Conference on Computer Vision and Pattern Recognition
  (CVPR)}, 770--778.

\bibitem[{iNaturalist(2019)}]{inaturalist19}
iNaturalist. 2019.
\newblock {iNaturalist} 2019 competition dataset.
\newblock ~\url{https://github.com/visipedia/inat_comp/tree/master/2019}.
\newblock Accessed: 2023-03-04.

\bibitem[{Ioffe and Szegedy(2015)}]{ioffe2015batch}
Ioffe, S.; and Szegedy, C. 2015.
\newblock Batch normalization: Accelerating deep network training by reducing
  internal covariate shift.
\newblock In \emph{International conference on machine learning}, 448--456.
  PMLR.

\bibitem[{Jabri, Owens, and Efros(2020)}]{jabri2020space}
Jabri, A.; Owens, A.; and Efros, A. 2020.
\newblock Space-time correspondence as a contrastive random walk.
\newblock \emph{Advances in neural information processing systems}, 33:
  19545--19560.

\bibitem[{Jenni and Favaro(2018)}]{jenni2018self}
Jenni, S.; and Favaro, P. 2018.
\newblock Self-supervised feature learning by learning to spot artifacts.
\newblock In \emph{Proceedings of the IEEE Conference on Computer Vision and
  Pattern Recognition}, 2733--2742.

\bibitem[{Krause et~al.(2013)Krause, Stark, Deng, and
  Fei-Fei}]{KrauseStarkDengFei-Fei_3DRR2013}
Krause, J.; Stark, M.; Deng, J.; and Fei-Fei, L. 2013.
\newblock 3D Object Representations for Fine-Grained Categorization.
\newblock In \emph{4th International IEEE Workshop on 3D Representation and
  Recognition (3dRR-13)}. Sydney, Australia.

\bibitem[{Krizhevsky(2009)}]{Krizhevsky2009LearningML}
Krizhevsky, A. 2009.
\newblock Learning multiple layers of features from tiny images.
\newblock Technical report, University of Toronto.

\bibitem[{Li et~al.(2021{\natexlab{a}})Li, Yang, Zhang, Gao, Xiao, Dai, Yuan,
  and Gao}]{li2021efficient}
Li, C.; Yang, J.; Zhang, P.; Gao, M.; Xiao, B.; Dai, X.; Yuan, L.; and Gao, J.
  2021{\natexlab{a}}.
\newblock Efficient self-supervised vision transformers for representation
  learning.
\newblock \emph{arXiv preprint arXiv:2106.09785}.

\bibitem[{Li et~al.(2021{\natexlab{b}})Li, Zhou, Zhang, Zhang, Wang, Jiang, Wu,
  and Wang}]{li2021dense}
Li, X.; Zhou, Y.; Zhang, Y.; Zhang, A.; Wang, W.; Jiang, N.; Wu, H.; and Wang,
  W. 2021{\natexlab{b}}.
\newblock Dense semantic contrast for self-supervised visual representation
  learning.
\newblock In \emph{Proceedings of the 29th ACM International Conference on
  Multimedia}, 1368--1376.

\bibitem[{Li et~al.(2021{\natexlab{c}})Li, Chen, Yang, Li, Zhu, Zhao, Deng, Wu,
  Zhao, Tang et~al.}]{li2021mst}
Li, Z.; Chen, Z.; Yang, F.; Li, W.; Zhu, Y.; Zhao, C.; Deng, R.; Wu, L.; Zhao,
  R.; Tang, M.; et~al. 2021{\natexlab{c}}.
\newblock Mst: Masked self-supervised transformer for visual representation.
\newblock \emph{Advances in Neural Information Processing Systems}, 34.

\bibitem[{Liu et~al.(2021{\natexlab{a}})Liu, Sangineto, Bi, Sebe, Lepri, and
  Nadai}]{liu2021efficient}
Liu, Y.; Sangineto, E.; Bi, W.; Sebe, N.; Lepri, B.; and Nadai, M.
  2021{\natexlab{a}}.
\newblock Efficient Training of Visual Transformers with Small Datasets.
\newblock \emph{Advances in Neural Information Processing Systems}, 34.

\bibitem[{Liu et~al.(2021{\natexlab{b}})Liu, Lin, Cao, Hu, Wei, Zhang, Lin, and
  Guo}]{liu2021swin}
Liu, Z.; Lin, Y.; Cao, Y.; Hu, H.; Wei, Y.; Zhang, Z.; Lin, S.; and Guo, B.
  2021{\natexlab{b}}.
\newblock Swin transformer: Hierarchical vision transformer using shifted
  windows.
\newblock In \emph{Proceedings of the IEEE/CVF International Conference on
  Computer Vision}, 10012--10022.

\bibitem[{Maji et~al.(2013)Maji, Rahtu, Kannala, Blaschko, and
  Vedaldi}]{maji13fine-grained}
Maji, S.; Rahtu, E.; Kannala, J.; Blaschko, M.~B.; and Vedaldi, A. 2013.
\newblock Fine-Grained Visual Classification of Aircraft.
\newblock \emph{ArXiv}, abs/1306.5151.

\bibitem[{Netzer et~al.(2011)Netzer, Wang, Coates, Bissacco, Wu, and
  Ng}]{Netzer2011}
Netzer, Y.; Wang, T.; Coates, A.; Bissacco, A.; Wu, B.; and Ng, A.~Y. 2011.
\newblock Reading Digits in Natural Images with Unsupervised Feature Learning.
\newblock In \emph{NIPS Workshop on Deep Learning and Unsupervised Feature
  Learning 2011}.

\bibitem[{Nilsback and Zisserman(2008)}]{Nilsback2008AutomatedFC}
Nilsback, M.-E.; and Zisserman, A. 2008.
\newblock Automated Flower Classification over a Large Number of Classes.
\newblock \emph{2008 Sixth Indian Conference on Computer Vision, Graphics \&
  Image Processing}, 722--729.

\bibitem[{Noroozi and Favaro(2016)}]{noroozi2016unsupervised}
Noroozi, M.; and Favaro, P. 2016.
\newblock Unsupervised learning of visual representations by solving jigsaw
  puzzles.
\newblock In \emph{European conference on computer vision}, 69--84. Springer.

\bibitem[{O~Pinheiro et~al.(2020)O~Pinheiro, Almahairi, Benmalek, Golemo, and
  Courville}]{o2020unsupervised}
O~Pinheiro, P.~O.; Almahairi, A.; Benmalek, R.; Golemo, F.; and Courville,
  A.~C. 2020.
\newblock Unsupervised learning of dense visual representations.
\newblock \emph{Advances in Neural Information Processing Systems}, 33:
  4489--4500.

\bibitem[{Oord, Li, and Vinyals(2018)}]{oord2018representation}
Oord, A. v.~d.; Li, Y.; and Vinyals, O. 2018.
\newblock Representation learning with contrastive predictive coding.
\newblock \emph{arXiv preprint arXiv:1807.03748}.

\bibitem[{Parkhi et~al.(2012)Parkhi, Vedaldi, Zisserman, and
  Jawahar}]{parkhi12a}
Parkhi, O.~M.; Vedaldi, A.; Zisserman, A.; and Jawahar, C.~V. 2012.
\newblock Cats and Dogs.
\newblock In \emph{IEEE Conference on Computer Vision and Pattern Recognition}.

\bibitem[{Pathak et~al.(2016)Pathak, Kr\"ahenb\"uhl, Donahue, Darrell, and
  Efros}]{pathakCVPR16context}
Pathak, D.; Kr\"ahenb\"uhl, P.; Donahue, J.; Darrell, T.; and Efros, A. 2016.
\newblock Context Encoders: Feature Learning by Inpainting.
\newblock In \emph{CVPR}.

\bibitem[{Pont-Tuset et~al.(2017)Pont-Tuset, Perazzi, Caelles, Arbel\'aez,
  Sorkine-Hornung, and {Van Gool}}]{Pont-Tuset_arXiv_2017}
Pont-Tuset, J.; Perazzi, F.; Caelles, S.; Arbel\'aez, P.; Sorkine-Hornung, A.;
  and {Van Gool}, L. 2017.
\newblock The 2017 DAVIS Challenge on Video Object Segmentation.
\newblock \emph{arXiv:1704.00675}.

\bibitem[{Tamkin et~al.(2021)Tamkin, Liu, Lu, Fein, Schultz, and
  Goodman}]{tamkin2021dabs}
Tamkin, A.; Liu, V.; Lu, R.; Fein, D.; Schultz, C.; and Goodman, N. 2021.
\newblock DABS: A Domain-Agnostic Benchmark for Self-Supervised Learning.
\newblock \emph{arXiv preprint arXiv:2111.12062}.

\bibitem[{Tartaglini, Vong, and Lake(2022)}]{tartaglini2022developmentally}
Tartaglini, A.~R.; Vong, W.~K.; and Lake, B.~M. 2022.
\newblock A Developmentally-Inspired Examination of Shape versus Texture Bias
  in Machines.
\newblock \emph{arXiv preprint arXiv:2202.08340}.

\bibitem[{Vaswani et~al.(2017)Vaswani, Shazeer, Parmar, Uszkoreit, Jones,
  Gomez, Kaiser, and Polosukhin}]{vaswani2017attention}
Vaswani, A.; Shazeer, N.; Parmar, N.; Uszkoreit, J.; Jones, L.; Gomez, A.~N.;
  Kaiser, {\L}.; and Polosukhin, I. 2017.
\newblock Attention is all you need.
\newblock \emph{Advances in neural information processing systems}, 30.

\bibitem[{Verma et~al.(2020)Verma, Kumawat, Nakashima, and
  Raman}]{verma2020yoga}
Verma, M.; Kumawat, S.; Nakashima, Y.; and Raman, S. 2020.
\newblock Yoga-82: a new dataset for fine-grained classification of human
  poses.
\newblock In \emph{Proceedings of the IEEE/CVF Conference on Computer Vision
  and Pattern Recognition Workshops}, 1038--1039.

\bibitem[{Wang et~al.(2021)Wang, Zhang, Shen, Kong, and Li}]{wang2021dense}
Wang, X.; Zhang, R.; Shen, C.; Kong, T.; and Li, L. 2021.
\newblock Dense contrastive learning for self-supervised visual pre-training.
\newblock In \emph{Proceedings of the IEEE/CVF Conference on Computer Vision
  and Pattern Recognition}, 3024--3033.

\bibitem[{Wu et~al.(2018{\natexlab{a}})Wu, Xiong, Yu, and
  Lin}]{wu2018unsupervised}
Wu, Z.; Xiong, Y.; Yu, S.~X.; and Lin, D. 2018{\natexlab{a}}.
\newblock Unsupervised feature learning via non-parametric instance
  discrimination.
\newblock In \emph{Proceedings of the IEEE Conference on Computer Vision and
  Pattern Recognition}, 3733--3742.

\bibitem[{Wu et~al.(2018{\natexlab{b}})Wu, Xiong, Yu, and
  Lin}]{Wu2018UnsupervisedFL}
Wu, Z.; Xiong, Y.; Yu, S.~X.; and Lin, D. 2018{\natexlab{b}}.
\newblock Unsupervised Feature Learning via Non-parametric Instance
  Discrimination.
\newblock \emph{2018 IEEE/CVF Conference on Computer Vision and Pattern
  Recognition}, 3733--3742.

\bibitem[{Xiao et~al.(2020)Xiao, Engstrom, Ilyas, and Madry}]{xiao2020noise}
Xiao, K.; Engstrom, L.; Ilyas, A.; and Madry, A. 2020.
\newblock Noise or signal: The role of image backgrounds in object recognition.
\newblock \emph{arXiv preprint arXiv:2006.09994}.

\bibitem[{Xiao et~al.(2018)Xiao, Liu, Zhou, Jiang, and Sun}]{xiao2018unified}
Xiao, T.; Liu, Y.; Zhou, B.; Jiang, Y.; and Sun, J. 2018.
\newblock Unified perceptual parsing for scene understanding.
\newblock In \emph{Proceedings of the European Conference on Computer Vision
  (ECCV)}, 418--434.

\bibitem[{Xiao et~al.(2021)Xiao, Reed, Wang, Keutzer, and
  Darrell}]{xiao2021region}
Xiao, T.; Reed, C.~J.; Wang, X.; Keutzer, K.; and Darrell, T. 2021.
\newblock Region similarity representation learning.
\newblock In \emph{Proceedings of the IEEE/CVF International Conference on
  Computer Vision}, 10539--10548.

\bibitem[{Xie et~al.(2021{\natexlab{a}})Xie, Lin, Yao, Zhang, Dai, Cao, and
  Hu}]{xie2021self}
Xie, Z.; Lin, Y.; Yao, Z.; Zhang, Z.; Dai, Q.; Cao, Y.; and Hu, H.
  2021{\natexlab{a}}.
\newblock Self-supervised learning with swin transformers.
\newblock \emph{arXiv preprint arXiv:2105.04553}.

\bibitem[{Xie et~al.(2021{\natexlab{b}})Xie, Lin, Zhang, Cao, Lin, and
  Hu}]{xie2021propagate}
Xie, Z.; Lin, Y.; Zhang, Z.; Cao, Y.; Lin, S.; and Hu, H. 2021{\natexlab{b}}.
\newblock Propagate yourself: Exploring pixel-level consistency for
  unsupervised visual representation learning.
\newblock In \emph{Proceedings of the IEEE/CVF Conference on Computer Vision
  and Pattern Recognition}, 16684--16693.

\bibitem[{Zhai et~al.(2022)Zhai, Jaitly, Ramapuram, Busbridge, Likhomanenko,
  Cheng, Talbott, Huang, Goh, and Susskind}]{Zhai2022PositionPA}
Zhai, S.; Jaitly, N.; Ramapuram, J.; Busbridge, D.; Likhomanenko, T.; Cheng,
  J.~Y.; Talbott, W.~A.; Huang, C.; Goh, H.; and Susskind, J.~M. 2022.
\newblock Position Prediction as an Effective Pretraining Strategy.
\newblock In \emph{ICML}.

\bibitem[{Zhang, Isola, and Efros(2016)}]{zhang2016colorful}
Zhang, R.; Isola, P.; and Efros, A.~A. 2016.
\newblock Colorful image colorization.
\newblock In \emph{European Conference on Computer Vision}, 649--666. Springer.

\bibitem[{Zhou et~al.(2017)Zhou, Zhao, Puig, Fidler, Barriuso, and
  Torralba}]{zhou2017scene}
Zhou, B.; Zhao, H.; Puig, X.; Fidler, S.; Barriuso, A.; and Torralba, A. 2017.
\newblock Scene parsing through ade20k dataset.
\newblock In \emph{Proceedings of the IEEE conference on computer vision and
  pattern recognition}, 633--641.

\bibitem[{Zhou et~al.(2021)Zhou, Wei, Wang, Shen, Xie, Yuille, and
  Kong}]{zhou2021ibot}
Zhou, J.; Wei, C.; Wang, H.; Shen, W.; Xie, C.; Yuille, A.; and Kong, T. 2021.
\newblock ibot: Image bert pre-training with online tokenizer.
\newblock \emph{arXiv preprint arXiv:2111.07832}.

\end{thebibliography}

\end{document}